\documentclass[10pt,journal,compsoc]{IEEEtran}
\ifCLASSINFOpdf
\else
\fi
\ifCLASSOPTIONcompsoc
\usepackage[nocompress]{cite}
\else
\usepackage{cite}
\fi
\ifCLASSINFOpdf
\else
\fi
\usepackage{algorithm}
\usepackage{setspace}

\usepackage{algpseudocode}
\usepackage{amsmath}

\usepackage{times}

\usepackage{amsfonts,amssymb} 

\usepackage{microtype}
\usepackage{graphicx}
\usepackage{epstopdf}
\usepackage{booktabs}
\usepackage{multirow}
\usepackage{verbatim}
\usepackage{url}

\usepackage{CJK}
\usepackage{algorithm}
\usepackage{algorithmicx}
\usepackage{algpseudocode}
\usepackage{amsmath}
\usepackage{bm}

\usepackage{threeparttable}
\usepackage{enumitem}
\floatname{algorithm}{algorithm}

\usepackage{cite}
\ifCLASSOPTIONcompsoc
	\usepackage[caption=false,font=normalsize,labelfont=sf,textfont=sf]{subfig}
\else
	\usepackage[caption=false,font=footnotesize]{subfig}
\fi

\begin{document}
%
\title{HYPER\(^2\): Hyperbolic Poincar\'e Embedding for Hyper-Relational Link Prediction}
%
%
%
%

\author{Shiyao~Yan,
        Zequn~Zhang*,
        Xian~Sun,~\IEEEmembership{Senior Member,~IEEE,}
    	Guangluan~Xu,~\IEEEmembership{Member,~IEEE,}
    	Li~Jin,
    	and~Shuchao~Li
\IEEEcompsocitemizethanks{\IEEEcompsocthanksitem S. Yan and X. Sun are with Aerospace Information Research Institute, Chinese Academy of Sciences, Beijing 100190, China, Key Laboratory of Network Information System Technology (NIST), Aerospace Information Research Institute, Chinese Academy of Sciences, Beijing 100190, China, University of Chinese Academy of Sciences, Beijing 100190, China and School of Electronic, Electrical and Communication Engineering, University of Chinese Academy of Sciences, Beijing 100190, China.  \protect\\
E-mail: yanshiyao19@mails.ucas.ac.cn, sunxian@mail.ie.ac.cn

\IEEEcompsocthanksitem  Z. Zhang, G. Xu, L. Jin and S. Li are with Aerospace Information Research Institute, Chinese Academy of Sciences, Beijing 100190, China and Key Laboratory of Network Information System Technology (NIST), Aerospace Information Research Institute, Chinese Academy of Sciences, Beijing 100190, China.

E-mail:\{zqzhang1,gluanxu,jinli,lishuchao\}@mail.ie.ac.cn


}

\thanks{Manuscript received April dd, 2020; revised mm dd, 2020.}}

%
%

\markboth{Journal of \LaTeX\ Class Files,~Vol.~14, No.~8, August~2015}%
{Shell \MakeLowercase{\textit{et al.}}:HYPER\(^2\):A Poincare Embedding method for n-ary Link Prediction}
%



\IEEEtitleabstractindextext{
\begin{abstract}
Link Prediction, addressing the issue of completing KGs with missing facts, has been broadly studied. However, less light is shed on the ubiquitous hyper-relational KGs. Most existing hyper-relational KG embedding models still tear an n-ary fact into smaller tuples, neglecting the indecomposability of some n-ary facts. While other frameworks work for certain arity facts only or ignore the significance of primary triple. In this paper, we represent an n-ary fact as a whole, simultaneously keeping the integrity of n-ary fact and maintaining the vital role that the primary triple plays. In addition, we generalize hyperbolic Poincar\'e embedding from binary to arbitrary arity data, which has not been studied yet. To tackle the weak expressiveness and high complexity issue, we propose HYPER\(^2\) which is qualified for capturing the interaction between entities within and beyond triple through information aggregation on the tangent space. Extensive experiments demonstrate HYPER\(^2\) achieves superior performance to its translational and deep analogues, improving SOTA by up to 34.5\% with relatively few dimensions. Moreover, we study the side effect of literals and we theoretically and experimentally compare the computational complexity of HYPER\(^2\) against several best performing baselines, HYPER\(^2\) is 49-61 times quicker than its counterparts.
\end{abstract}

}

\maketitle

\IEEEdisplaynontitleabstractindextext

%
\IEEEpeerreviewmaketitle

\IEEEraisesectionheading{\section{Introduction}\label{sec:introduction}}

%
%
%
%
\IEEEPARstart{T}{he} emerging of large-scale Knowledge Graphs(KGs), such as Freebase\cite{freebase}, Wikidata\cite{wikidata} and Google's Knowledge Graph\cite{Google} power a plethora of downstream tasks ranging from recommendation systems\cite{2016Collaborative} to semantic search\cite{2017Explicit} and from question answering\cite{2015Semantic} to natural language understanding\cite{2020ERNIE1}. KGs are generally represented as a set of triplets in the form of \emph{(head entity, relation, tail entity)} where a relation links an entity pair. Despite containing rich information, KGs are still in face of incompleteness issue. To our knowledge, 71\% of all people in Freebase have no $place\_of\_birth$ attribute\cite{2015A}. Link prediction task, seeking to predict new links between entities in the graph based on existing ones, tackles this issue\cite{2017TKDE}.

A variety of representitive link prediction techniques such as RESCAL\cite{2011RESCAL}, TransE\cite{2013TransE} and ConvE\cite{2017ConvE} have been devised to work over triplet based KGs. And in the past decade, we have witnessed gratifying progress in this realm. Nevertheless, when lots of efforts are poured into binary facts(i.e. triplets), we pay less attention to the ubiquitous n-ary facts involving more than two entities. For instance, \emph{award\_nomination} relation often involves an awarding institution, a receiver, an award and a work. And several performers can also co-occur in a performance.
As a matter of fact, in Freebase, more than 1/3 entities participate in n-ary facts\cite{2016On}. These higher-arity facts with more knowledge are much closer to natural language, link prediction in hyper-relational KGs provides an excellent potential for lots of downstream NLP applications\cite{2018HighLife}. Existing triple based models decompose n-ary facts into a set of triplets, irreversibly resulting in internal structural and semantic loss\cite{2020HINGE}. 

\begin{figure}[!t]
	\centering
	\subfloat[]{\includegraphics[width=1.5in]{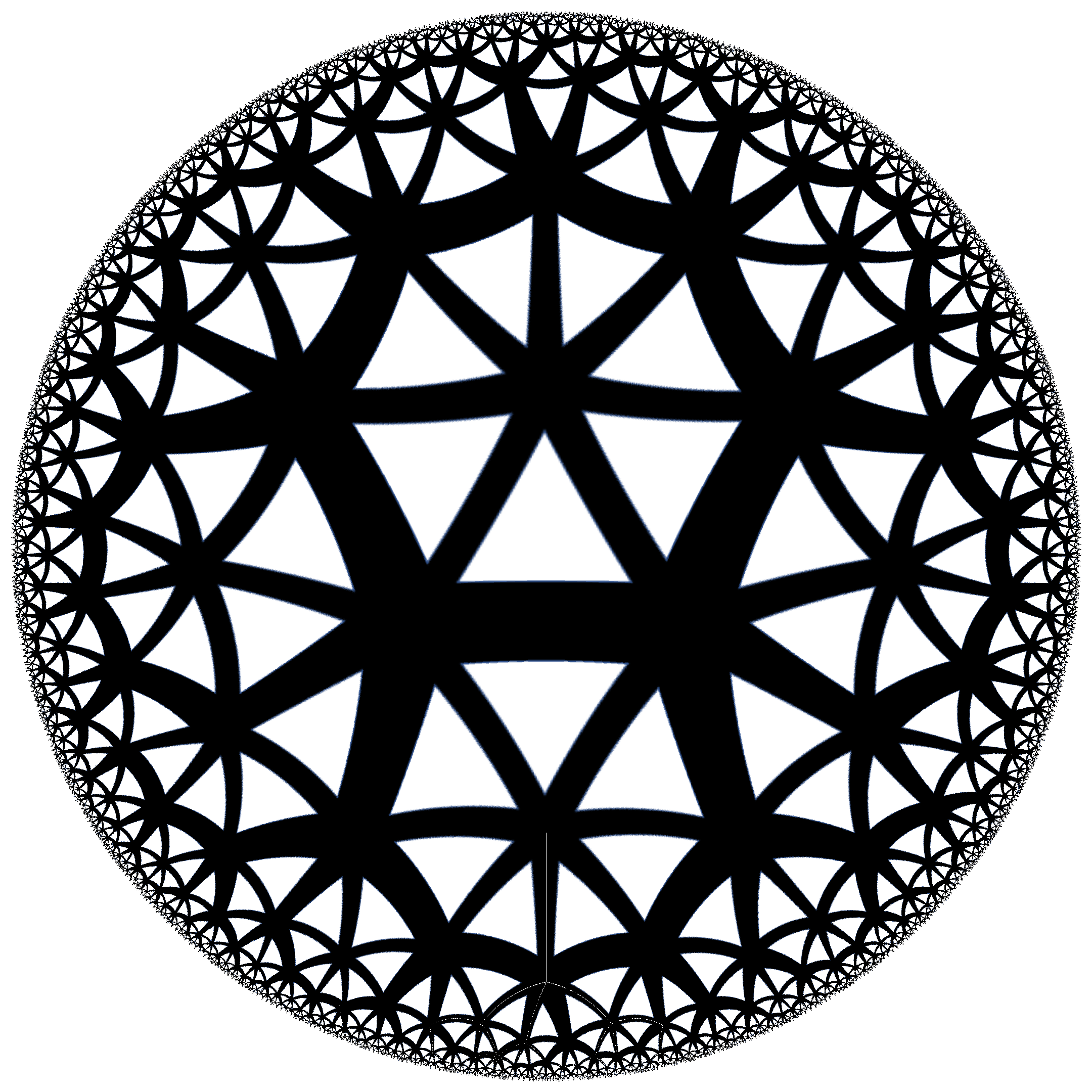}
		\label{fig_first_case}}
	\hfil
	\subfloat[]{\includegraphics[width=1.5in]{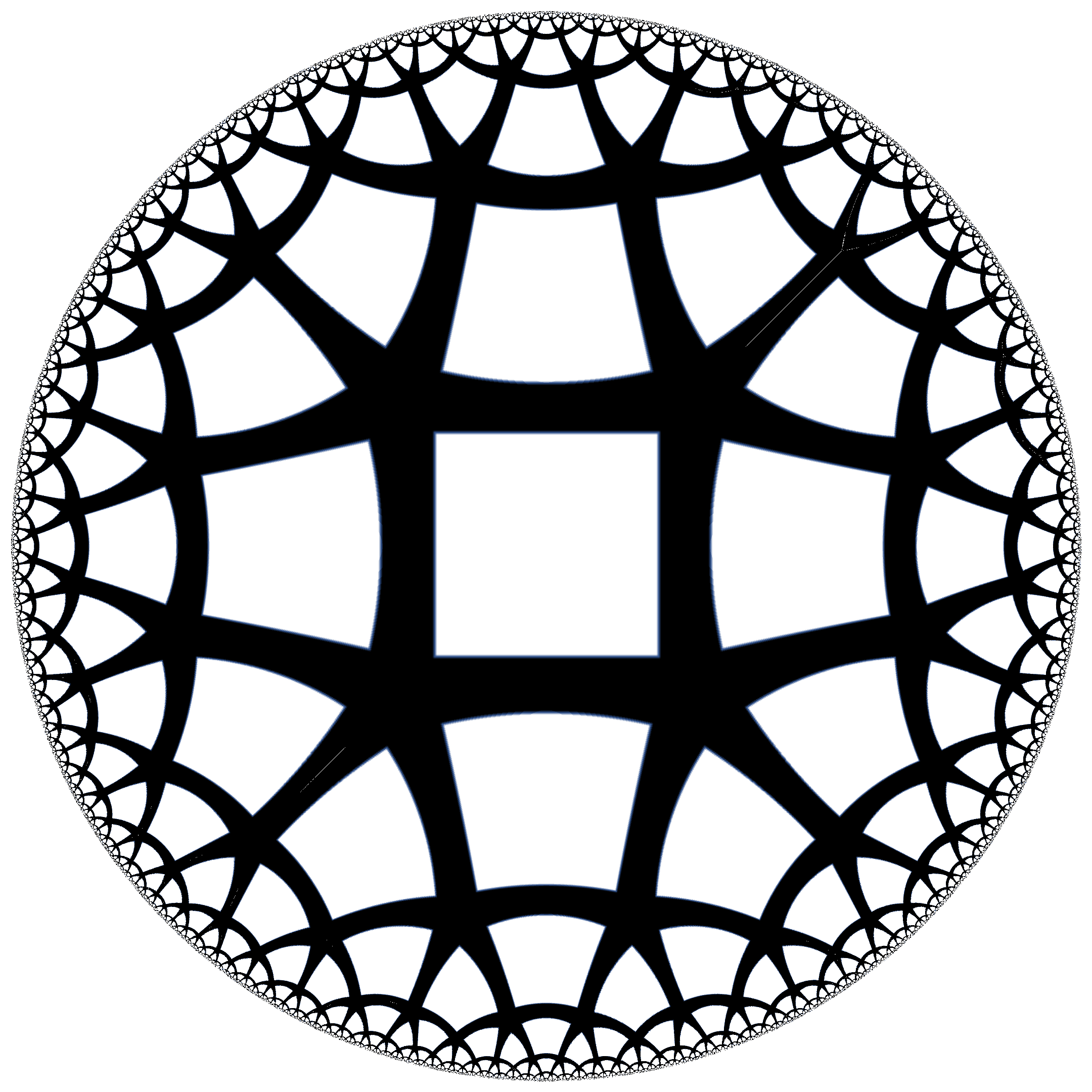}
		\label{fig_second_case}}
	\caption{ (a) is a Poincar\'e ball with 3 vertices in each polygon and 7 polygons adjacent to each vertex. (b) is a Poincar\'e ball with 4 vertices in each polygon and 5 polygons adjacent to each vertex. As shown, a Poincar\'e ball contains rich geometric structure.}
	\label{figure1}
\end{figure}

\begin{figure*}[!t] 
	\centering
	
	\includegraphics[width=18.5cm,height=4.5cm]{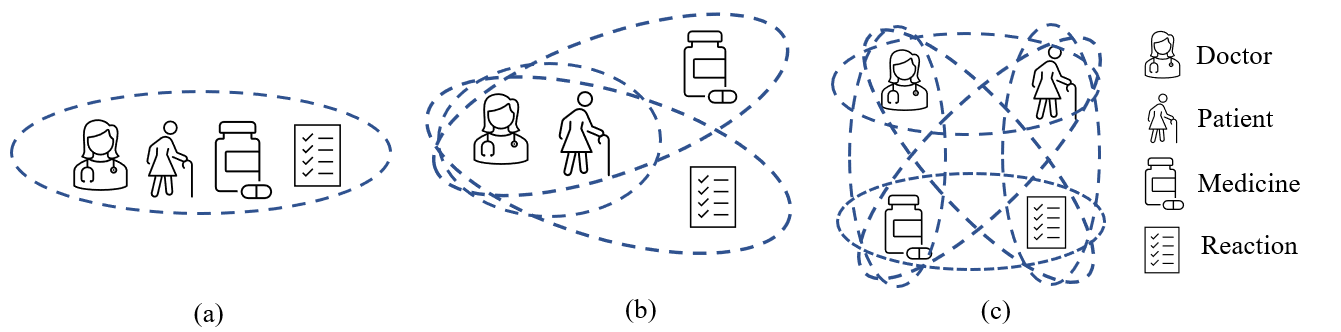}
	\caption{Take the medical\_record relation as an instance, (a) shows that our proposed HYPER\(^2\) keeps the integrity of an n-ary fact, (b) demonstrate that HINGE views an n-ary fact as a primary triple and $n-2$ 3-ary tuples. Reaction of a patient is closely related to the medicine he takes, it's not reasonable to seperate reaction from medicine. (c) illustrates that NaLP calculates relatedness between each entity pairs. It greatly undermines the structrue of an n-ary fact and it can't be scalable to large-scale hyper-relational KGs. }
	\label{illustrate} 
\end{figure*}

To deal with the more challenging n-ary facts without breaking them into triplets, m-TransH\cite{2016On}, a translational distance model based on TransH is first proposed. Along with the model, the authors put forward the first hyper-relational KG JF17K, which is derived from Freebase with frequent presence of n-ary facts. The work is ground breaking, but it combines relations and entities in sequence so as to get a so-called meta-relation. The strict position restriction would bring information loss, for in some cases, two or more entities changing position doesn't necessarily lead to semantic deviation. Take the sentence \emph{Andrea Bocelli sang Time To Say Goodbye and Canto Della Terra with Sarah Brightman in the concert Vivere: Live In Tuscany} as an example, exchanging position of \emph{Time To Say Goodbye} and \emph{Canto Della Terra} expresses exactly the same semantic meaning. RAE\cite{2018RAE} further takes multi-layer perceptron into m-TransH, the incremental work brings constant progress. But both of the two tranlational models are limited by the translation nature and suffer from weak expressiveness,
leaving much room for improvement. Besides, these two models cannot predict relations.

 Deep models recently gain popularity in n-ary link prediction just as they prevail in many other tasks. NaLP and a new dataset WikiPeople, extracted from Wikidata mainly focusing on people-related facts, are proposed in \cite{2019NaLP}. NaLP outperforms m-TransH and RAE on both dataset, but as shown in Figure \ref{illustrate}(c), it breaks an n-ary fact into $n$ key-value pairs and needs to compute relatedness $C_n^2$ times for a single fact. Authors of HINGE\cite{2020HINGE} advocate that primary triple keeps the main sturcture and semantic information of an n-ary fact.
 Experiments shows that Hinge greatly outperforms NaLP, validating the significance of primary triple. But as dipicted in Figure \ref{illustrate}(b), HINGE views an n-ary fact as the combiantion of a primary triple and $n-2$ 3-ary tuples, it captures triple and quintuple features with convolution network. It has been proved in \cite{2019hyperpath} that some n-ary facts are in possesseion of indecomposibility. Take the $medical\_record$ relation as an example, A $doctor$ will record the $reaction$ of a $patient$ after he takes $medicine$. $Reaction$ of a $patient$ is tightly correlated to what $medicine$ he takes, it's not reasonable to seperate $reaction$ from $medicine$. Despite the remarkable performance improvement, n-ary facts in these deep models still suffer from non-integrity issue. Moreover, deep models always sustain relatively high computational complexity.

HSimplE\cite{2020hype} and HyPE\cite{2020hype} are convolutional models, but they can only deal with 2-ary to 6-ary data. The tensor decomposition based model GTED\cite{2020GTED} also sustains inflexibility issue, it's only applicable to 3-ary and 4-ary facts. Moreover, the three models don't explicitly distinguish entities within and beyond triple, neglecting the importance of the primary triple. In this work, we represent an n-ary fact as a whole, keeping the integrity of the fact and maintain the vital role that the primary triple plays in an n-ary fact.

Recently, advance in binary link prediction task illustrates that hyperbolic Poincar\'e model shows superiority and lower complexity over its counterparts\cite{2019Murp}. We further learn that Poincar\'e ball excels in representing knowledge with hierarchy for the exponential growth of distance on the boundary of the hyperbolic space. As shown in Figure 1, containing rich geometric structures, a Poincar\'e ball is endowed with strong expressiveness in representing complex graph-structured data. It's also notable that hyperbolic Poincar\'e embedding method has an edge in complexity in that it involves much less
matrix addition and multiplication operations compared to existing deep models. To our konwledge, it has not been studied in hyper-relational KG.


Considering the expressiveness and low complexity, we deem that hyperbolic Poincar\'e embedding method, which to the best of our knowledge has not been studied in hyper-relational KGs yet, is promising for a both efficient and effective model. In this work, we propose a generalized hyperbolic Poincar\'e embedding framework that is appilicable to arbitary arity facts, termed as HYPER\(^2\).
 In HYPER\(^2\), all entities are first randomly initialized on the hyperbolic space, then through vector projection between hyperbolic space and tangent space and aggregation of information on the tangent space, HYPER\(^2\) captures the interaction between the primary triple and the affiliated entities. 

HYPER\(^2\) achieves SOTA on two representative hyper-relational datasets, improving HITS@1, HITS@10, MRR by 34.5\%, 21.9\%, 29.1\% on WikiPeople, and 23.1\%, 14.4\%, 18.9\% on JF17K respectively, which validates the effectiveness and superiority of our model. Besides, HYPER\(^2\) is less computationally complicated, 49-61 times quicker in comparison with its analogues. Moreover,
 our experiments verify literals (numeric values, date-time instances or other string) impede embedding learning. Our contribution can be summarized as follows:

\begin{itemize}[leftmargin=0.5cm]
	\item HYPER\(^2\) employs hyper-relation to totally shrink from tearing an n-ary facts into smaller tuples, simultaneously considering the vital role that the primary triple plays in an n-ary fact and the indecomposibility of some n-ary facts. 
	
	\item To the best of our knowledge, our proposed HYPER\(^2\) is the first model that employs hyperbolic Poincar\'e ball for hyper-relational KGs,  HYPER\(^2\) generalize hyperbolic Poincar\'e method from binary to arbitary arity facts, illustrating high flexibility.
	
	\item Extensive experiments on two representative hyper-relational KGs demonstrate that HYPER\(^2\) realizes superior performance to its counterparts, improving SOTA by a surprisingly large margin, demonstrating absolute superiority.
	
	\item We validate the side effect brought by literals in Wikipeople dataset. And we theoretically and experimentally compare the parameter size and the computational complexity of  HYPER\(^2\) against several previous best-performing baselines. Experiments show our model greatly overwhelms its counterparts, unveiling low complexity.

\end{itemize}


\section{Related Work}


\subsection{Graph Embedding Approaches in non-Euclidean Geometry}
There has been a growing number of work in embedding graph data in non-Euclidean spaces. Hyperbolic embedding was brought into link prediction in the lexical database WordNet on the Poincar\'e ball in\cite{2017poincare} by Nickel and Kiela, who show that hyperbolic embedding with low dimension is capable of significantly outperforming their higher-dimensional Euclidean counterparts in terms of both generalization ability and representation capability. Subsequently, the same team embeds hierarchical data in the Lorentz model of hyperbolic geometry\cite{2018Lorentz}. Combining hyperbolic geometry with deep learning, Hyperbolic Neural Networks\cite{2018HNN}, Hyperbolic Attention Networks\cite{2019HAN} and Hyperbolic Graph Convolution NetWork\cite{2019HGCN} are proposed. More recently, it is noted that non-uniform hierarchy\cite{2019product} like circular structrue can be found in data so that graph starts to be embedded on a product manifold with components of different curvature(spherical, hyperbolic and Euclidean). Following Poincar\'e GloVe\cite{2019pglove}, another multi-relational Poincar\'e based method Murp\cite{2019Murp}, employing bilinear scoring function and bias for head and tail entity, shows strong expressive power in modelling hierarhy. In \cite{2020rotate-poin}, hyperbolic Poincar\'e embedding is combined with RotatE\cite{2019RotatE}. To the best of our knowledge, embedding  hyper-relational KGs on a hyperbolic poincar\'e ball is unexplored. 

\subsection{Binary Link Prediction}

Grouped by scoring function, link prediction models in binary relational KGs can be roughly classified into three categories: translational distance models, neural network models, and semantic matching models. 

Translational distance models calculate the entity distance after a translational operation carried out by the relation. TransE\cite{2013TransE} is the earliest translational distance model, 
but it fails to represent 1-N ,N-1, N-N relation. Since TransE, a variety of variants such as TransH\cite{2014TransH}, TransR\cite{2017TransR}, TransMS\cite{2019TransMs} have been proposed to overcome the problem. 

With the prosperity of deep learning, a variety of deep models are developed. 
ConvKB\cite{2018ConvKB}, ConvE\cite{2017ConvE} utilize convolution network. RSN\cite{2019RSN} learns to exploit long-term relational dependencies in knowledge graphs with residual learning. R-GCN\cite{2018RGCN} and GAE\cite{2019GAE} apply graph convolutional network to model KGs. 
 Facilitated with various neural network structures, these models achieve superior results\cite{2020Asurvey}. 

Semantic matching models like RESCAL\cite{2011RESCAL}, DistMult\cite{2015distmult}, ComplEx\cite{2016Complex} and TuckER\cite{2019tucker} are bilinear frameworks. RotatE\cite{2019RotatE} treats relation as a rotation from head entity to tail entity. 
QuatE\cite{2019Quaternion} extends embedding from complex space to hypercomplex space. 
HAKE\cite{2020HAKE} maps entities into the polar coordinate system to model semantic hierarchies. 
 Murp\cite{2019Murp} maps entities into a poincar\'e ball in the hyperbolic space to capture the semantic hierarchies. Compared to neural network-based models, bilinear ones achieve competitive results but are more interpretable.

\subsection{Hyper-Relational Link Prediction}
 Ways to predict new links in the more challenging hyper-relational KGs can be roughly divided into translational distance and neural network based models. 
 

 In respect to translational distance models, m-TransH\cite{2016On} is the first work to cope with hyper-relational KGs. 
m-TransH directly extends TransH from binary to n-ary facts, 
As the ground breaking work in n-ary link prediction, experiment on JF17K dataset shows that m-TranH exceeds TransH by a cheerful margin. RAE\cite{2018RAE} further intakes multi-layer perceptron into m-TransH. 
 The translation nature of the two models incapacitates them from fully expressing the more complicated n-ary relations \cite{2018simple}.
 
 NaLP, HSimplE\cite{2020hype}, HypE\cite{2020hype}, NaLP-fix\cite{2020HINGE}, HINGE, etc. are deep models. 
 NaLP outperforms m-TransH and RAE on both JF17K and WiKipeople. However, in NaLP, an n-ary fact is torn into n key-value pairs. Akin to the triple-based decomposition, such pair-wise breakdown could also greatly undermine the structure of an n-ary fact.
   HINGE views an n-ary fact as a primary triple and $n-2$ quintuples, capturing triple-wise features and quintuple-wise features with convolution neural network. 
 Authors of Hinge also proposed a variant of NaLP, i.e. NaLP-fix, where the model structure stays the same, but a new negative sampling strategy is given. 
 HSimplE and HypE devise a set of convolution kernels for facts with different arity, suffering from flexibility. 
 GTED \cite{2020GTED} extends Tucker to n-ary case, i.e., n-Tucker, and further integrates Tensor Ring decomposition 
 into n-Tucker. 
 GTED is not flexible, working for 3-ary and 4-ary data only. 
 Besides, entities within and beyond triple are not differentiated explicitly.
 
  Our HYPER\(^2\) considers both the vital role that the primary triple plays and the integrity of an n-ary fact. Moreover, taking the advantage of the Poincar\'e ball, HYPER\(^2\) requires fewer parameters.
 
 \section{Preliminary}
 
 \subsection{Basic Concepts}
 Hyper-relational knowledge graph contains both binary facts and n-ary facts, we take entities within and beyond primary triple as primary entities and affiliated entities, respectively. Link prediction tasks in hyper-relational KGs includes head/tail entity prediction, relation prediction and affilated entity prediction. We will give their definitions below.
 \newtheorem{definition}{Definition}
 \begin{definition}
  \textbf{Hyper-Relational Knowledge Graph}. Given a hyper-relational KG \emph{G} with a set of relations \emph{R} and a set of entities \emph{E}, a fact in hyper-relational KG contains a relation and arbitary number of entities. Formally, a fact can be written as a tuple $F:(r,{e_1},{e_2},...{e_i},...,{e_{n}})$, where $r \in R$, $e_i \in E$. $n$ stands for the number of entities participating in the tuple $F$. If $n=2$, $F$ is a \textbf{binary fact}. If $n > 2$, $F$ is an \textbf{n-ary fact} and $r$ is a \textbf{hyper-relation}, which differs from an edge in hypergraph for containing label and direction infromation. We take $(r,{e_1},{e_2})$ as \textbf{primary triple}. Entities within primary triple, i.e. $e_1, e_2$, are taken as \textbf{primary entities}, entities beyond primary triples, i.e. entities in \{$e_i |i \in (2,n], i \in \mathbb{Z}$\} are denoted as \textbf{affiliated entities}. 
 \end{definition}
 
 Take the sentence that \emph{James Horner was nominated for the 68th Academy Awards for Best Original Dramatic Score for his excellent work in BraveHeart} as an instance. The fact implied in the sentence can be written as \emph{(be\_nominated\_for, James Horner, the 68th Academy Awards for Best Original Dramatic Score, BraveHeart)} where \emph{James Horner} is head entity, \emph{the 68th Academy Awards for Best Original Dramatic Score} is tail entity, these two entities are all primary entities. While \emph{BraveHeart} is the affiliated entity and $be\_nominated\_for$ is the hyper-relation.
 
 \begin{definition}
 	\textbf{Hyper-Relational Link Prediction Task}. Suppose \emph{G} is an imcomplete hyper-relational KG, hyper-relational link prediction task aims at inferring missing facts on the basis of known facts in \emph{G}. In practice, the problem is simplified to predict the missing entity or relation for a partial fact. In this work, predicting primary entity for partial fact $(r,?,{e_2},...,{e_i},...,{e_{n}})$ or $(r,{e_1},?,...,{e_i},...,{e_{n}})$
 	is denoted as \emph{head/tail entity prediction}, inferring the missing entity for $(r,{e_1},{e_2},...,?,...,{e_{n}})$ is termed as \emph{affiliated entity prediction}. While predicting missing relation$(?,{e_1},{e_2},...,{e_i},...,{e_{n}})$ is named as \emph{relation prediction}.
 \end{definition}

 \subsection{Hyperbolic geometry of the Poinca\'e ball} 
\begin{figure}[!t] 
	\centering
	
	\includegraphics[width=7cm]{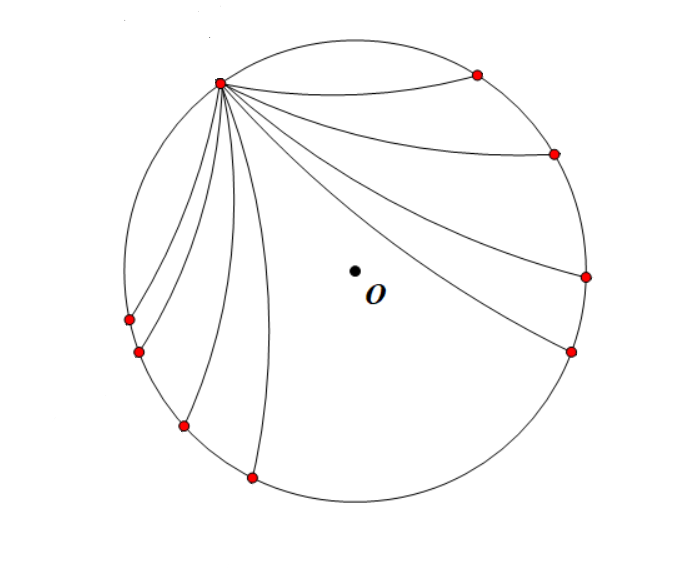}
	\caption{shows the shortest path between two points on a Poincare ball. }
	\label{figure3} 
\end{figure}

An d-dimensional Poincar\'e ball $(\mathbb{P}_K^d,{g^\mathbb{P}})$ with radius $\frac{1}{{\sqrt K }}(K> 0)$   is a real and smooth manifold $\mathbb{P}_K^d = \left\{ {{\mathbf{x}} \in {\mathbb{R}^d}:K{{\left\| {\mathbf{x}} \right\|}^2} < 1} \right\}$ where ${g^\mathbb{P}}$ is a Riemannian metric conformal to the Euclidean metric ${g^\mathbb{E}} = {{\mathbf{I}}_d}$. And we have ${g^\mathbb{P}} = {\left( {\lambda _{\mathbf{x}}^K} \right)^2}{g^\mathbb{E}}$  where $ \lambda _{\mathbf{x}}^K = 2/(1 - {\left\| {\mathbf{x}} \right\|^2})$ is an conformal coefficient. For each point ${\mathbf{x}} \in \mathbb{P}_K^d$, the metric tensor ${g^\mathbb{P}}$ defines a positive-definite inner product ${\mathcal{T}_{\mathbf{x}}}\mathbb{P}_K^d \times {\mathcal{T}_{\mathbf{x}}}\mathbb{P}_K^d \to \mathbb{R}$, where  $\mathcal{T}_{\mathbf{x}}\mathbb{P}_K^d$ is an d-dimensional Euclidean space tangential to $\mathbb{P}_K^d$ at $\mathbf{x}$. To project a point ${\mathbf{x}} \in \mathbb{P}_K^d$ to its corresponding tangent space $\mathcal{T}_{\mathbf{x}}\mathbb{P}_K^d$, there exists a logarithm map $\log _{\mathbf{x}}^{\text{K}}:\mathbb{P}_K^d \to {T_{\mathbf{x}}}\mathbb{P}_K^d$. Inverse is a exponential map $\exp _{\mathbf{x}}^K:{T_{\mathbf{x}}}\mathbb{P}_K^d \to \mathbb{P}_K^d$. For a Poincar\'e ball, they are:
\begin{equation}
\exp _{\mathbf{x}}^K({\mathbf{v}}) = {\mathbf{x}}{ \oplus _K}\left( {\tanh \left( {\sqrt K \frac{{\lambda _{\mathbf{x}}^K\left\| {\mathbf{v}} \right\|}}{2}} \right)\frac{{\mathbf{v}}}{{\sqrt K \left\| {\mathbf{v}} \right\|}}} \right),\label{equation1}
\end{equation}

\begin{equation}
\log _{\mathbf{x}}^K({\mathbf{v}}) = \frac{2}{{\sqrt K \lambda _{\mathbf{x}}^K}}{\tanh ^{ - 1}}\left( {\sqrt K \left\| { - {\mathbf{x}}{ \oplus _K}{\mathbf{v}}} \right\|} \right)\frac{{ - {\mathbf{x}}{ \oplus _K}{\mathbf{v}}}}{{\left\| { - {\mathbf{x}}{ \oplus _K}{\mathbf{v}}} \right\|}}\label{equation2}
\end{equation}
where ${\mathbf{v}} \in {T_{\mathbf{x}}}\mathbb{P}_K^d$ is a vector tangential to $\mathbb{P}_K^d$ at $\mathbf{x}$, $\left\|  \bullet  \right\|$ represents the Euclidean norm and ${ \oplus _K}$ denotes M\"obius addition\cite{2001MADD}:
\begin{equation}
{\mathbf{x}}{ \oplus _K}{\mathbf{y}} = \frac{{\left( {1 + 2K\langle {\mathbf{x}},{\mathbf{y}}\rangle  + K{{\left\| {\mathbf{y}} \right\|}^2}} \right){\mathbf{x}} + \left( {1 - K{{\left\| {\mathbf{x}} \right\|}^2}} \right){\mathbf{y}}}}{{1 + 2K({\mathbf{x}},{\mathbf{y}}) + {K^2}{{\left\| {\mathbf{x}} \right\|}^2}{{\left\| {\mathbf{y}} \right\|}^2}}}.\label{equation3}
\end{equation}
$\left\langle { \bullet , \bullet } \right\rangle $ stands for Euclidean inner product. To look into the formulas, exponential (logarithm) map mainly involves non-linear hyperbolic tangent (inverse hyperbolic tangent) operation, division, norm and M\"obius addition, while in M\"obius addition, scalar-vector multiplication, Euclidean norm and inner product are involved. Compared to the large quantities of calculation involvod in convolutional layer and fully connected network, these seemingly complex formulas actually take less time to compute. We will theoretically and experimentally analyze the computational complexity of several deep model in Section 5.
Matrix-vector multiplication also has a M\"obius counterpart\cite{2018HNN}:
\begin{equation}
{\mathbf{M}}{ \otimes _K}{\mathbf{x}} = \exp _0^K\left( {{\mathbf{M}}log_0^K({\mathbf{x}})} \right)\label{equation4}.
\end{equation}
In \eqref{equation4}, a point ${\mathbf{x}} \in \mathbb{P}_K^d$ is first projected onto the Euclidean tangent space at ${\mathbf{0}} \in \mathbb{P}_K^d$ with logarithm map, then multiplied by a matrix ${\mathbf{M}} \in {\mathbb{R}^{d \times d}}$, finally projected back to hyperbolic space $\mathbb{P}_K^d$ with exponential map. The matrix $\mathbf{M}$ in our model is diagonal, further reducing computational complexity.
The distance between two points ${\mathbf{x}},{\mathbf{y}} \in \mathbb{P}_K^d$ is given by the length of geodesic. Geodesic defines the shortest path between two points that lies in a given surface. In Euclidean space, geodesic is the line segment that links two endpoints, while for Poincar\'e ball, the geodesic is depicted in Figure \ref{figure3}, length of geodesic is given by:
\begin{equation}
{d_\mathbb{P}}({\mathbf{x}},{\mathbf{y}}) = \frac{2}{{\sqrt K }}{\tanh ^{ - 1}}\left( {\sqrt K \left\| { - {\mathbf{x}}{ \oplus _K}{\mathbf{y}}} \right\|} \right).
\end{equation}

Calculating the length of geodesic is not onerous as well. Along with the introduction of hyperbolic geometry of Poincar\'e ball, it's also demonstrated that Poincar\'e ball is superior to its counterparts in terms of computational complexity. In the section to come, we will introduce more details about our methodology.


\section{Hyperbolic Poincar\'e Embedding for Hyper-Relational Link Prediction}
Most graph-structured data possesses inherent geometric structures, hyper-relational KG is no exception. Among these geometric structures, tree-like hierarchical structure is universal. Hyperbolic Poincar\'e embedding method, which to our knowledge is adept in representing hierarchy and has not been studied in hyper-relational KGs, is promising for an effective and efficient model for n-ary link prediction. We propose HYPER\(^2\) to embed arbitary ariry facts in hyper-relational KG on a hyperbolic Poincar\'e ball.

\begin{figure*}[!t] 
	\centering
	
	\includegraphics[width=18cm,height=7.5cm]{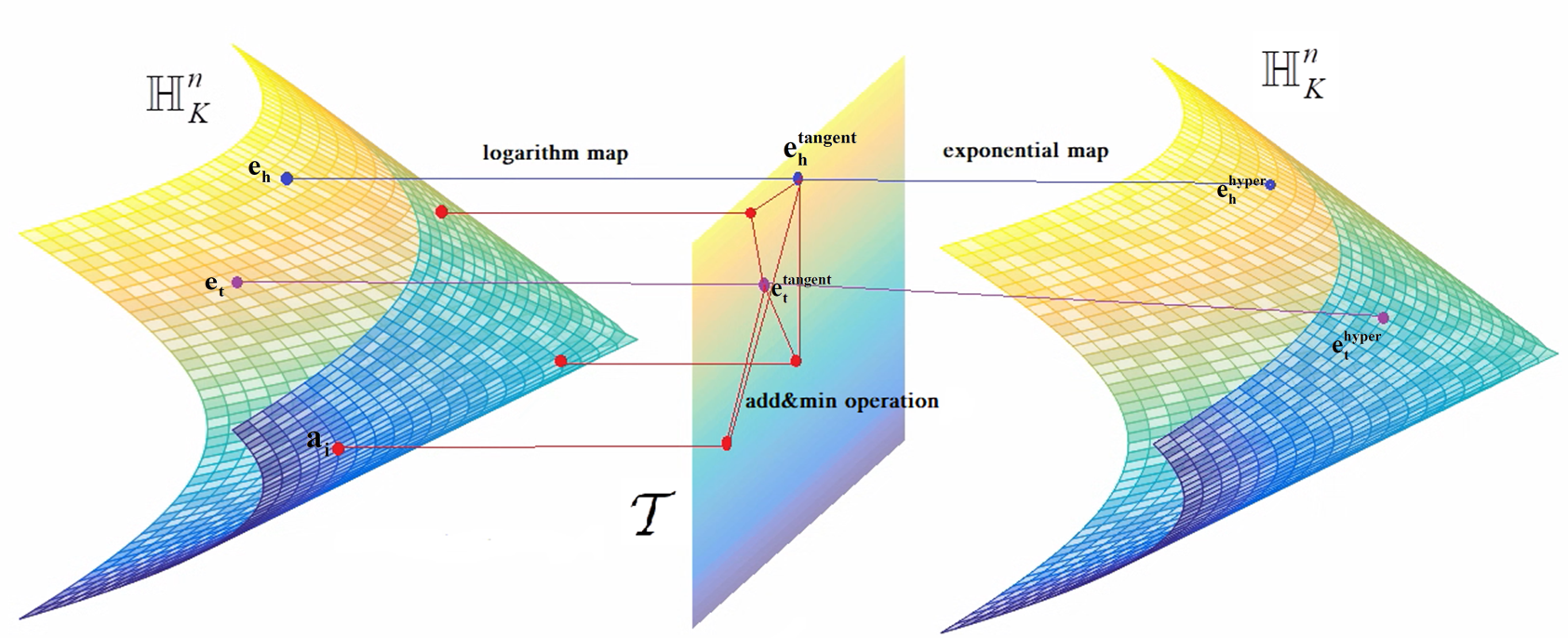}
	\caption{illustrates structure of HYPER\(^2\). All entities are first initialized on hyperbolic space and then projected to the tangent space where addition and element-wise minimization operation are performed to capture interaction of primary(head and tail) entities and affiliated entites. Next, primary entities with rich information are projected back to hyperbolic space for final scoring. }
	\label{model} 
\end{figure*}

\subsection{Affiliated Information Specific Entity Representation}
Fact in JF17K is in the form of $(r, e_1, e_2,...,e_n)$, while fact in WikiPeople is organized as $(e_h,r,e_t,r_1,a_1,...r_i,a_i,...)$. Since HYPER\(^2\) takes the relation in n-ary fact as a hyper-relation, we remove $r_i$ in WikiPeople and Wiki-filtered, and leaves the hyper-relation $r$ only. Facts in the three datasets are uniformly represented as $(r, e_h, e_t, a_1,...,a_i,...a_2)$. The structure of HYPER\(^2\) is shown in Figure \ref{model}.

In consideration of the significance of the primary triple, ${e_h},{e_t}$ are treated as primary entities, $a_i, i \in (2,n-2]$ is taken as affiliated entity. In spite of the significance of the primary triple, the rich information in affiliated entities is nonnegligible. It's salient to capture the interaction between the primary triple and affiliated entities. With this consideration, we design an information aggregator to draw affiliated information specific entity representation.

Detailly, in our model, entities and a hyper-relation in an n-ary fact $F:(r,{e_h},{e_t},{a_1},...,{a_i},...,{a_{n - 2}})$ are first initialized as d-dimensional vectors ${\mathbf{r}},{{\mathbf{e}}_h},{{\mathbf{e}}_t}, \cdots ,{{\mathbf{a}}_i}, \cdots ,{{\mathbf{a}}_{n - 2}}$ on a Poincar\'e ball $\mathbb{P}_K^d$, then all entities are projected to tangent Euclidean space with a logarithm map. In order to model those useful information introduced by the dynamic interaction between the primary triple and affiliated entities, we add latent vector of each affiliated entity ${\mathbf{a}}_i$ to head entity and tail entity respectively:
\begin{equation}
	{\mathbf{e}}_h^i = log_0^K({{\mathbf{e}}_h}) + log_0^K({{\mathbf{a}}_i}),
	\label{equation6}
\end{equation}
\begin{equation}
	{\mathbf{e}}_t^i = log_0^K({{\mathbf{e}}_t}) + log_0^K({{\mathbf{a}}_i}).
	\label{equation7}
\end{equation}
Subsequently, we concatenate the embeddings with affiliated information and perform an element-wise minimization operation:
\begin{equation}
{\mathbf{e}}_h^{tangent} = \min (concat({\mathbf{e}}_h^1, \cdots ,{\mathbf{e}}_h^{{\text{n}} - 2})),
\label{equation8}
\end{equation}
\begin{equation}
{\mathbf{e}}_t^{tangent} = \min (concat({\mathbf{e}}_t^1, \cdots ,{\mathbf{e}}_t^{{\text{n}} - 2})).
\label{equation9}
\end{equation}
We utilize the minimization operation with the assumption that the framework telling right from wrong with least affiliated information is a better model. we aim to 1) judge a valid n-ary fact as true with only scarce presence of affiliated information. 2) judge an invalid fact as false without affiliated information in abundance. In similar manner, previous works NaLP and HINGE have successfully merged the relatedness vector derived from neural networks. Lastly, the affiliated inforamtion specific entity representation, i.e. ${\mathbf{e}}_h^{tangent},{\mathbf{e}}_t^{tangent}$, are projected back to hyperbolic space with exponential map:
\begin{equation}
{\mathbf{e}}_h^{hyper} = \exp _0^K({\mathbf{e}}_h^{tangent})\label{equation10},
\end{equation}
\begin{equation}
{\mathbf{e}}_t^{hyper} = \exp _0^K({\mathbf{e}}_t^{tangent})\label{equation11}.
\end{equation}

When n=2, i.e. no affiliated entity participates in a fact, our model can generalize to the binary Poincar\'e embedding model Murp. In HYPER\(^2\), all entities are initialized on hyperbolic space and then projected to tangent space where we capture the interaction and aggregate the information of primary and affiliated entities via an element-wise minimization and an addition operator. Then those entities with aggregated information are projected back to hyperbolic space. Next, we will give the scoring function of evaluating a fact.

\subsection{Scoring Function}
Scoring function of bilinear models contains a bilinear product between the head entity embedding, a relation matrix and the object entity embedding. Effectiveness of bilinear models have been verified in \cite{2011RESCAL}\cite{2019tucker}. Nevertheless, there doesn't exist a clear replacement of the Euclidean inner product in hyperbolic spaces. On the basis of the Poincar\'e Glove\cite{2019pglove},  we model the inner product with a distance function, i.e. $\langle {\mathbf{x}},{\mathbf{y}}\rangle  = \frac{1}{2}( - d{({\mathbf{x}},{\mathbf{y}})^2} + {\left\| {\mathbf{x}} \right\|^2} + {\left\| {\mathbf{y}} \right\|^2})$, and we take the place of the squared norm with ${b_h}$ and ${b_t}$. Additionally, enlightened by \cite{2019Murp}, which applys relation-specific transformations to head entity and tail entity respectively, i.e. a stretch by a diagonal predicate matrix $\bm{R_h} \in {\mathbb{R}^{d \times d}}$ to head entity and a translation by a d-dimensional vector offset ${\mathbf{r}}$ to tail entity, a basic scoring function to score a fact $F$ can be given as:
	\begin{equation}
	\phi (F) =  - d{(\bm{R_h}{{\mathbf{e}}_h},{{\mathbf{e}}_t} + {\mathbf{r}})^2} + {b_h} + {b_t}.\label{equation12}
	\end{equation}
In our HYPER\(^2\), the information of affiliated entities have been aggregated to the primary entities, so the scoring function can reflect both the relatedness of the primary triple and the relatedness of the whole fact. Take the hyperbolic analogue of \eqref{equation12}, we can define the scoring function of HYPER\(^2\) as:

\begin{equation}
\begin{aligned}
\phi (F) &= \phi (r,{e_h},{e_t}, \cdots ,{a_i}, \cdots ,{a_{n - 2}}) \hfill \\
{\text{        }} &=  - {d_\mathbb{P}}{(\bm{R_h}{ \otimes _K}{\mathbf{e}}_h^{hyper},{\mathbf{e}}_t^{hyper}{ \oplus _K}{\mathbf{r}})^2} + {b_h} + {b_t} \hfill \\
{\text{        }} &=  - {d_\mathbb{P}}{(\exp _0^K(\bm{R_h}\log _0^K({\mathbf{e}}_h^{hyper})),{\mathbf{e}}_t^{hyper}{ \oplus _K}{\mathbf{r}})^2}    \hfill \\
{\text{         }} &{\text{     }} \ \ \ \ + {b_h} + {b_t} \label{equation13}
\end{aligned}
\end{equation}
where we replace Euclidean addition and matrix-vector multiplication with their m\"obius equivalents. And we replace the head and tail entity representation with their affiliated information specific form. HYPER\(^2\) is applicable to facts with arbitary arity, sharing high complexity. In case of no affiliated entity, HYPER\(^2\) can generalize to Murp, retaining the ability to model triples.
\subsection{Training and Optimization}
As a commonly used data augmentation method in literature\cite{2019Murp,2020Diachronic}, we bring reciprocal relation $({r^{ - 1}},{e_t},{e_h}, \cdots ,{a_i}, \cdots ,{a_{n - 2}})$ for each  fact $F:(r,{e_h},{e_t},...,{a_1},...,{a_{n - 2}})$ in dataset. And we generate $nneg$ negative samples for each true fact in training set by randomly corrupt an entity domain or the relation domain. Specifically, we corrupt the relation with another randomly chosen relation from the relation set \emph{R}. While for entity, we corrupt head entity $(r,{e_h}',{e_t}, \cdots ,{a_i}, \cdots ,{a_{n - 2}})$, tail entity $(r,{e_h},{e_t}', \cdots ,{a_i}, \cdots ,{a_{n - 2}})$ or one of the affiliated entity $(r,{e_h},{e_t}, \cdots ,{a_i}', \cdots ,{a_{n - 2}})$ with a randomly selected entity from the whole entity set \emph{E}. Models are trained to minimize the binary cross-entropy loss:
\begin{equation}
\mathcal{L}=  - \frac{1}{N}\sum\limits_{m = 1}^N {\left( {{y_m}\log \left( {{p_m}} \right) + \left( {1 - {y_m}} \right)\log \left( {1 - {p_m}} \right)} \right)} \label{equation14}
\end{equation}
where $N$ stands for the number of training samples, $y_m$ is a binary label indicating whether a fact is positive or not, ${p_m} = \sigma (\phi (F))$ represents the predicted probability, and $\sigma ( \bullet )$ is a sigmoid function. We optimize the model with Riemannian stochastic gradient descent (RSGD)\cite{2013RSGD}. There exist two main differences between RSGD and SGD. One is the computation of gradient, Riemannian gradient ${\nabla _R}\mathcal{L}$ is obtained by multiplying the Euclidean gradient ${\nabla _E}\mathcal{L}$ with the inverse of the Poincar\'e metric tensor i.e.,${\nabla _R}\mathcal{L} = \frac{1}{{{{\left( {\lambda _{\mathbf{x}}^K} \right)}^2}}}{\nabla _E}\mathcal{L}$, Another difference lies in the update step, the Euclidean update step is:
\begin{equation}
 \theta  \leftarrow \theta  - \eta {\nabla _E}\mathcal{L} \label{equation15},
\end{equation}
while the Riemannian update step is:
\begin{equation}
\theta  \leftarrow \exp _{\mathbf{x}}^K\left( { - \eta {\nabla _R}\mathcal{L}} \right) \label{equation16}
\end{equation} The training process of HYPER\(^2\) is give in algorithm 1.
\begin{algorithm}
	\caption{Training Process of HYPER\(^2\)}
		\begin{algorithmic}[1] 
				\Require TrainingSet $T$, EntitySet $E$, RelationSet $R$, max number of epochs $nepoch$, Embedding Dimensions $d$, batch size $\beta$, lerning rate $\eta$, number of negative samples $nneg$
				\Ensure Relation Embedding $\bm{M_R}$, Entity Embeddings $\bm{M_E}$, Bias for Head ${b_h}$, Bias for Tail ${b_t}$ and Diagonal Weight Matrix $\bm{W}$
				\State Initialnize  $\bm{M_R,M_E} \gets 10^{-3}*random\_norm(0,1)$ 
				\Statex \qquad \qquad    $\bm{W} \gets uniform(-1, 1)$
				\Statex \qquad \qquad    ${b_h,b_t} \gets zero\_initializer()$
				\Repeat
				\For{$i = 1 \to\left[ {\frac{{\left| T \right|}}{\beta }} \right] $}
				\State $mini\_batch \gets sample(T , \beta)$
				\State $\mathcal{L} \gets 0$ 
				\For {$F+  \in mini\_batch$}
				\State $F- \gets negative\_sample(T , nneg)$
				\State $F \gets F+  \cup F- $
				\Statex \qquad \qquad // look up the embedding of involved entities
				\State $\bm{F_E} \gets look\_up(\bm{M_E}, F_E)$ 
				\Statex \qquad \qquad // look up the embedding of involved relations
				\State $\bm{F_R} \gets look\_up(\bm{M_R}, F_R)$ 
				\State $S({F}) \gets$ evaluate F following \eqref{equation13}
				\State $\mathcal{L}(F) \gets$compute loss following \eqref{equation14}
				\State $\mathcal{L} \gets \mathcal{L}+\mathcal{L}(F)$
				\EndFor
				\State Update $\bm{M_E}$, $\bm{M_R}$ using RSGD following \eqref{equation15}
				\State Update $b_x$, $b_y$ and $\bm{W}$ using SGD following (\ref{equation16})  
				\EndFor
				\Until the evaluation results on the validation set drop continuously or this process has been iterated for nepoch times
		\end{algorithmic}
\end{algorithm}
In Section 4, we thoroughly introduce how HYPER\(^2\) works. In the coming Section 5, we will elaborate the experiment we carry out, state our observation and analyze our results.

\section{Experiments}
In this section, we will present our experimental setup, report experimental results and discuss our observations. Detailly, we will evaluate our proposed model HYPER\(^2\) on various n-ary link prediction tasks and do case study. Additionally, we will theoretically and experimentally compare the space and time complexity of our proposed HYPER\(^2\) against the main-stream deep models. Moreover, we will analyze the side effect that literals in Wikipeople bring about.

\begin{table*}[htbp]
	\centering
	\caption{Statistics of Dataset}
	\resizebox{18cm}{!}{
	\begin{tabular}{llllllllllll}
		\toprule[1.25pt]
		\multirow{2}{*}{Dataset}                              &\multirow{2}{*}{\#E}                    & \multirow{2}{*}{\#R}                  & \multicolumn{3}{c}{Training Set} & \multicolumn{3}{c}{Validation Set} & \multicolumn{3}{c}{Testing Set} \\
		\cmidrule{4-12}
		&                        &                      & 2ary     & Nary      & overall   & 2ary      & nary      & overall    & 2ary     & nary     & overall   \\
		\midrule
		\multirow{2}{*}{JF17K}               & \multirow{2}{*}{28645} & \multirow{2}{*}{322} & 44210    & 32169     & 76379     & -         & -         & -          & 10417    & 14151    & 24568     \\
		&                        &                      & 57.9\%   & 42.1\%   & 100\%     & -         & -         & -          & 42.4\%   & 57.6\%   & 100\%     \\
		\midrule
		\multirow{2}{*}{Wikipeople}          & \multirow{2}{*}{47765} & \multirow{2}{*}{193} & 270179   & 35546     & 305725    & 33845     & 4378      & 38223      & 33890    & 4391     & 38281     \\
		&                        &                      & 88.4\%   & 11.6\%    & 100\%     & 88.5\%    & 11.5\%    & 100\%      & 88.5\%   & 11.5\%   & 100\%    \\
		\midrule
		\multirow{2}{*}{Wikipeople-flitered} & \multirow{2}{*}{34825} & \multirow{2}{*}{144} & 280497   & 7397      & 287894    & 36597     & 977       & 37574      & 36627    & 971      & 37597     \\
		&                        &                      & 97.4\%   & 2.6\%     & 100\%     & 97.4\%    & 2.6\%    & 100\%      & 97.4\%   & 2.6\%   & 100\% \\
		\bottomrule[1.25pt]
		\label{table1}
	\end{tabular}}
\end{table*}

\subsection{Experiment Setup}
\subsubsection{Dataset}
We experiment on two hyper-relational datasets JF17K and WikiPeople, which are two commonly used datasets composed of both binary facts and n-ary facts. JF17K is derived from Freebase and has a high proportion of n-ary facts, while WikiPeople is filterd from Wikidata and holds a less frequent presence of n-ary facts. As the original WikiPeople dataset contains literals in some facts\cite{2020HINGE}, we remove those binary facts with a literal tail entity. While for n-ary facts, we reformat the original fact by removing all literals in the fact. The filtered dataset is termed as Wiki-filtered. Table \ref{table1} shows statistics of the datasets.

As observed from table \ref{table1}, Wiki-flitered contains only 2.6\% n-ary facts after removing those facts with literals. Despite the flaw, we use Wiki-filtered in our experiments. 

\subsubsection{Metrics}
Link prediction is a classic task for Knowledge Graph Completion. Given two domains of a binary fact or given two domains in the primary triple of an n-ary fact, the task is to predict the missing domain, i.e., we predict head entity $(r,?,e_t)$ or tail entity $(r,e_h,?)$, where the question mark denotes the missing domian. We introduce the protocols to evaluate a model by taking predicting missing tail entity $(r,e_h,?)$ as an instance. For the primary triple $(r,e_h,?)$ in one test (n-ary) fact, the missing tail entity is replaced by all entities in $E$, generating a group of candidate facts. Among the candidates, apart from the fact to be tested, other corrupted facts in the training/validate/test sets may also be true facts; these leaking facts are cancelled while the rest facts in candidate set are sent to the model to be scored. We get a ranking list ordered by the score and the rank of the testing fact is recorded. After evaluation for all facts in test set, we report Hits@1, Hits@3, Hits@10 and Mean Reciprocal Rank (MRR), which are metrics in wide use. The metrics and protocol for evaluating tail entity are also applied to predicting missing head entity $(r,?,e_t)$. We calculate the average results of head and tail entity prediction and denote the two tasks as a whole, i.e, head/tail entity prediction. Mertics and protocals for relation and affiliated entity prediction is more or less the same, lengthy details will not be given anymore.

\subsubsection{Baselines}
We compare HYPER\(^2\) against a collection of translational distance and neural network based models designed for link prediction on hyper-relational KGs. We cite the results of m-TransH, RAE and HyPE. It's noted that only a quarter of the facts in test set are evaluated in \cite{2020HINGE}, so extra experiments are conducted on more recently proposed model NaLP, NaLP-fix and HINGE. 
\begin{table*}[htbp]
	\centering
	\caption{Overall Results of Link Prediction on Head/Tail Entity}
	\begin{tabular}{lllllllllllll}
		\toprule[1.25pt]
		\multirow{2}{*}{Method} & \multicolumn{4}{c}{JF17K}     & \multicolumn{4}{c}{WikiPeople} & \multicolumn{4}{c}{Wiki-filtered} \\ \cmidrule{2-13}
		& MRR   & H@10  & H@3   & H@1   & MRR     & H@10   & H@3    & H@1    & MRR     & H@10    & H@3    & H@1      \\ \midrule
		
		m-TransH                & 0.206 & 0.463 & -     & 0.206 & -       & -      & -      & -      & 0.063   & 0.301   & -      & 0.063    \\
		RAE                     & 0.215 & 0.469 & -     & 0.215 & -       & -      & -      & -      & 0.059   & 0.306   & -      & 0.059    \\
		NaLP                    & 0.313 & 0.464 & 0.334 & 0.237 & 0.340   & 0.457  & 0.366  & 0.276  & 0.343   & 0.464   & 0.372  & 0.280    \\
		NaLP-fix                    & 0.346 & 0.502 & - & 0.259 & 0.345   & 0.473  & -  & 0.281  & 0.348   & 0.469   & -  & 0.281     \\
		HyPE                   & - &- & -     & - & -   & -  & -      & -  & -   & -   & -      & -     \\
		HINGE                   & \underline{0.490} &\underline{ 0.652} & -     & \underline{0.406} & \underline{0.351}   & \underline{0.480}  & -      & \underline{0.284}  & \underline{0.356}   & \underline{0.482}   & -      & \underline{0.289}     \\
		HYPER\(^2\)                  & \textbf{0.583} & \textbf{0.746} & \textbf{0.620} & \textbf{0.500} & \textbf{0.453}   & \textbf{0.585}  & \textbf{0.485}  & \textbf{0.382}  & \textbf{0.461}   & \textbf{0.597}   & \textbf{0.492}  & \textbf{0.391}   \\
		\bottomrule[1.25pt] \label{table2}
	\end{tabular}
\end{table*}

\begin{table*}[htbp]
	\centering
	\caption{Fine-grained Results of Head/Tail Prediction}
	
	\begin{tabular}{llllllllllllll}
		\toprule[1.5pt]
				\multirow{2}{*}{Fact Type} & \multirow{2}{*}{Method} & \multicolumn{4}{c}{JF17K}     & \multicolumn{4}{c}{WikiPeople} & \multicolumn{4}{c}{Wiki-filtered} \\ \cmidrule(l){3-14} 
		&                         & MRR   & H@10  & H@3   & H@1   & MRR     & H@10   & H@3    & H@1    & MRR      & H@10     & H@3     & H@1     \\ \cmidrule(r){1-14}
		\multirow{4}{*}{Bianry}    & NaLP                    & 0.118 & 0.246 & 0.121 & 0.058 & 0.351   & 0.465  & 0.374  & 0.291  & 0.348    & 0.464    & 0.372   & 0.287   \\
		& NaLP-fix                & 0.149 & 0.284 & - & 0.085 & 0.358  & 0.484  & 0.385  & 0.297  & \underline{0.358}    & 0.476    & 0.389   & \underline{0.293}   \\
		& HyPE               & - & 0.466 & - & - & -   & -  & -  & -  &-    & -    & -   & -   \\
		& HINGE                   & \underline{0.305} & \underline{0.507} & -     & \underline{0.207} & \underline{0.365}   & \underline{0.490}  & -      & \underline{0.299}  & \underline{0.358}    & \underline{0.483}    & -       & 0.292   \\
		& HYPER\(^2\)                  & \textbf{0.395} & \textbf{0.599} & \textbf{0.427} & \textbf{0.295} & \textbf{0.461}   & \textbf{0.589}  & \textbf{0.493}  & \textbf{0.394}  & \textbf{0.462}    & \textbf{0.597}    & \textbf{0.492}   & \textbf{0.391}   \\
		\hline
		\midrule
		\multirow{4}{*}{N-ary}     & NaLP                    & 0.456 & 0.625 & 0.491 & 0.369 & 0.237   & 0.384  & 0.262  & 0.161  & 0.261    & 0.411    & 0.286   & 0.187   \\
		& NaLP-fix                & 0.491 & 0.662 & - & 0.387 & \underline{0.245}   & 0.388  & -  & \underline{0.164}  & 0.265    & 0.418    & -   & \underline{0.193}   \\
		& HyPE               & - & - & - & - & -   & -  & -  & -  &-    & -    & -   & -   \\
		& HINGE                   & \underline{0.627} & \underline{0.759} & -     & \underline{0.554} & 0.244   & \underline{0.406}  & -      & 0.162  & \underline{0.267}    & \underline{0.430}    & -       & 0.185   \\
		& HYPER\(^2\)                  & \textbf{0.721} & \textbf{0.856} & \textbf{0.762} & \textbf{0.651} & \textbf{0.359}   & \textbf{0.518}  & \textbf{0.394}  & \textbf{0.275}  & \textbf{0.391}    & \textbf{0.570}    & \textbf{0.417}   & \textbf{0.304}   \\ 
		
		\bottomrule[1.25pt] \label{table3}
	\end{tabular}
\end{table*}

\subsubsection{Implemention Detail}
Our model is implemented in PyTorch and hyper-parameters are chosen via grid search. Concretely speaking, we select the learning rate $\eta$ from $\{1, 10, 30, 50, 80, 100\}$, the batch size $\beta$ from $\{64, 128, 256\}$, the number of negative samples $nneg$ from $\{25, 50, 100\}$. We find that the optimal $\eta$, $\beta$ and $nneg$ is $30, 128, 100$ for JF17K. While for WikiPeople and Wiki-filterd the best $\eta$, $\beta$ and $nneg$ is $80, 128, 100$. For the latter two datasets, we tune the model by MRR on the validate set. But for lack of validate set in JF17K, we tune the model on test set directly. As for other hyperparameters, we set the curvature of Poincar\'e ball to $K = 1$ and dimensions of embedding to $d = 50$. We run $400$ epochs for WikiPeople and WikiPeople-filterd while $800$ epochs for JF17K according to their respective convergence rate.

\subsection{Head/Tail Entity Prediction and Impact of Literals}
We report both coarse-grained and fine-grained results of head/tail entity prediction in this part. This experiment is conducted on JF17K, WikiPeople and Wiki-filtered, the impact of literals is analyzed along with the analysis of the experimental results.

\subsubsection{Overall Results of Head/Tail Prediction}
We calculate the average value of head and tail entity for each metric, our experimental results against the chosen baselines are shown in Table \ref{table2}. The best results are in boldface and the second best results are underlined.

As observed from Table 2, our proposed model HYPER\(^2\) constantly outperforms all baselines on all metrics by a surprisingly large margin. Among the baselines, HINGE shows better performance than others (i.e.,m-TransH, RAE, NaLP and NaLP-fix). Compared with the best baseline HINGE on Wiki-filtered, in terms of H@1, HYPER\(^2\) achieves a tremendous improvement of 35.2\% (34.5\% on WikiPeople and 23.2\% on JF17K, respectively). And our improvement on other metrics is not marginal as well,  HYPER\(^2\) increases H@10 on Wiki-filtered by 23.9\% (21.9\% on WikiPeople and 14.4\% on JF17K) and yield a significant improvement of 29.5\% on MRR (29.1\% on WikiPeople and 19.0\% on JF17K). The results verify the superior expressiveness of HYPER\(^2\) over its tranlational and deep counterparts.

It's also observed from the results of the two Wikipeople related datasets that the involvement of literals doesn't bring expected drastic performance drop. One feasible explanation can be that binary facts are in dominant majority in both datasets, accounting for as much as 88.4\% and 97.6\% respectively, thus n-ary facts can contribute very little to the overall results. To look into the impact of literals, we will further analyze fine-grained results of head/tail prediction.

\subsubsection{Fine-grained Results of Head/Tail Prediction}
Due to the subpar performance of m-TransH and RAE, we no longer report them in the following experiment. Results of HYPER\(^2\) and its counterparts on binary and n-ary facts are listed in Table \ref{table3} where the best results are highlighted and the second best results are underlined.

As recorded in table 3, our proposed HYPER\(^2\) substantially outperforms all baselines on all metrics by a large margin in both binary and n-ary cases. In terms of MRR on binary facts,  HYPER\(^2\) surpasses the second best HINGE by 29.5\%, 29.1\%, 26.3\% on Wiki-filtered, WikiPeople and JF17K, respectively. While the number is 46.4\%, 47.1\%, 15.0\% for n-ary facts. The cheerful fine-grained results illustrate that HYPER\(^2\) has strong power to model both binary and n-ary facts, validating that our model can well embed both the primary triple and the integral fact.

In bianry cases, all model shows similar performance on WikiPeople and Wiki-filtered, the binary results are more or less the same as the overall results in Table 2, which can be explained by the overwhelming dominance of binary facts as well. However, things are quite different for those n-ary facts with literals involved. We observe that the filtering of literals in WikiPeople brings obvious performance gain for head/tail entity prediction. It is after filtering literals that HYPER\(^2\) elevates MRR by 8.9\% from 0.359 to 0.391 and HINGE improves MRR by 9.4\% from 0.244 to 0.267, while HYPER\(^2\) elevates Hits@1 by 10.5\% to 0.304 and HINGE improves Hits@1 by 14.2\% to 0.185. The improvement brought the removal of literals indicates that literals in WikiPeople impede the learning a better embedding. For this reason, when we conduct the following experiments, we cancel WikiPeople and experiment on Wiki-filtered and JF17K only.

\subsection{Relation Prediction}

HyPE doesn't perform relation prediction and the model works for 2-ary to 6-ary data only, which means it can't be implemented on Wikipeople. For these reasons, HyPE is not reported anymore. As observed in Table \ref{table4}, Experiment shows superiority of HYPER\(^2\) in modelling relation, too. The baseline model HINGE has shown relatively good performance on relation prediction task, leaving very little room for further substantial improvement. But in terms of MRR, HYPER\(^2\) still overwhelm HINGE with 5.8\% and 1.5\% performance gain on JF17K and Wiki-filtered, respectively. It's noted that on JF17K, HYPER\(^2\) achieves much more significant improvement on n-ary facts than on bianry facts, one feasible explanation could be the expressiveness of HYPER\(^2\) in modelling n-ary relations. 
In terms of Hits@10, HYPER\(^2\) realizes superior or competitive results. But in terms of Hits@1, HYPER\(^2\) sees a definite excess on both JF17K and Wiki-filtered, demonstrating the power of HYPER\(^2\) to more accurately predict a missing relation.

\subsection{Affiliated Entity Prediction}

Similar to the previous experiment, experiment is conducted on Wiki-filtered and JF17K, and HYPER\(^2\) is compared with NaLP, NaLP-fix and HINGE only. Table \ref{table5} shows the results of affiliated entity prediction. It's shown that HYPER\(^2\) constantly outperforms HINGE, surpassing HINGE by 8.3\%, 7.5\%, 10.6\% in terms of MRR, Hits@10 and Hits@1 on JF17K dataset. While on Wiki-filtered, HYPER\(^2\) sees a competitive Hits@1 and MRR, but yeilds a 7.2\% improvement with regard to Hits@10. The performance gain on the two benchmark datasets shows that the affiliated infromation is effectively learned by the information aggregator designed on the tagent space.

\begin{table*}[htbp]
	
	\caption{Results of Relation Prediction}
	\label{table4}
	\begin{tabular}{lllllllllllllc}
		\toprule[1.25pt]
		\multirow{2}{*}{Dataset}            & \multirow{2}{*}{Method} & \multicolumn{3}{c}{MRR} & \multicolumn{3}{c}{Hits@10} & \multicolumn{3}{c}{Hits@3} & \multicolumn{3}{c}{Hits@1} \\
		\cmidrule{3-14}
		&                         & 2ary  & nary  & overall & 2ary   & nary   & overall   & 2ary   & nary   & overall  & 2ary   & nary   & overall  \\ \cmidrule{1-14}
		
		\multirow{4}{*}{JF17K} 
		& NaLP                    &0.811       &0.831       &0.825         &0.928        &0.927        &0.927           &0.872        &0.874        &0.873          &0.738        &0.773        &0.762          \\
		& NaLP-fix                &0.841       &0.862       &0.853         &0.943        & \underline{0.940 }      &0.942           &-        &-        & -         &0.759        & 0.798       &  0.781        \\
		& HINGE                   &\underline{0.898}       &\underline{0.898}       &\underline{0.898}         & \textbf{0.985}       &0.912        & \underline{0.943}          &  -      & -       & -         & \underline{0.833}       &   \underline{0.889 }    &  \underline{0.866  }      \\
		& HYPER\(^2\)                  & \textbf{0.929 }   & \textbf{0.966}      & \textbf{0.950}        &  \underline{0.975}      & \textbf{0.977}       &  \textbf{0.976}         &  \textbf{0.957}      & \textbf{0.969}       & \textbf{0.964}         & \textbf{0.898}       & \textbf{0.959}       & \textbf{0.933}    \\  \hline \cmidrule{1-14}
		\multirow{4}{*}{Wiki-filtered}             
		& NaLP                    & 0.738      &0.831       & 0.740        &0.941        &0.922        & 0.940          &  0.856      & 0.835       &0.855          &0.586        &0.670        &0.588          \\
		& NaLP-fix                & 0.757      &0.847       & 0.758        &0.953        &  0.930      &0.952           & -       &-        & -         & 0.601       & 0.692       &0.602          \\
		& HINGE                   &\underline{0.932}       & \underline{0.945}      
		&\underline{0.933}
		& \textbf{0.997}       &\textbf{0.998}        & \textbf{0.997}          & -       & -       &  -        &   \underline{0.887}     & \underline{0.898}       & \underline{0.887}         \\
		& HYPER\(^2\)                  &\textbf{0.947}       & \textbf{0.948}      & \textbf{0.947}        &\underline{0.994}        &  \underline{0.994}      & \underline{0.987}          &  \textbf{0.980}      &  \textbf{0.977}      & \textbf{0.980}         &\textbf{0.914}        & \textbf{0.918}       & \textbf{0.914}\\
		\bottomrule[1.25pt]        
	\end{tabular}
\end{table*}
\begin{table*}[htbp]
	\centering
	\caption{Results of Affiliated Entity Prediction}
	\label{table5}
	\begin{tabular}{lcccccccc}
		\toprule[1.25pt]
		\multirow{2}{*}{Method} & \multicolumn{4}{c}{JF17K}     & \multicolumn{4}{c}{Wiki-filtered} \\
		\cmidrule{2-9}
		& MRR   & Hits@10  & Hits@3   &  Hits@1   	& MRR   & Hits@10  & Hits@3   &  Hits@1       \\
		\cmidrule{1-9}
		NaLP    	& 0.510 & 0.655 &0.545  & 0.432 & 0.375   & 0.602   &0.413 & 0.244    \\
		NaLP-FIX   & 0.532 & 0.679  &-  & 0.687  & 0.443   & 0.616   &-  &0.265    \\
		HINGE     & \underline{0.575} & \underline{0.696}   &-  & \underline{0.501} & \underline{0.502}   & \underline{0.643}     &-   & \textbf{0.428}    \\
		HYPER\(^2\)    & \textbf{0.623} & \textbf{0.748}  &\textbf{0.660}   & \textbf{0.554} & \textbf{0.507}   & \textbf{0.689}    &\textbf{0.543}   & 0.423    \\
		\bottomrule[1.25pt]
	\end{tabular}
	
\end{table*}

\begin{figure*}[!t]
	\centering
	\subfloat[Visualization of HYPER\(^2\)]{\includegraphics[width=3.3in]{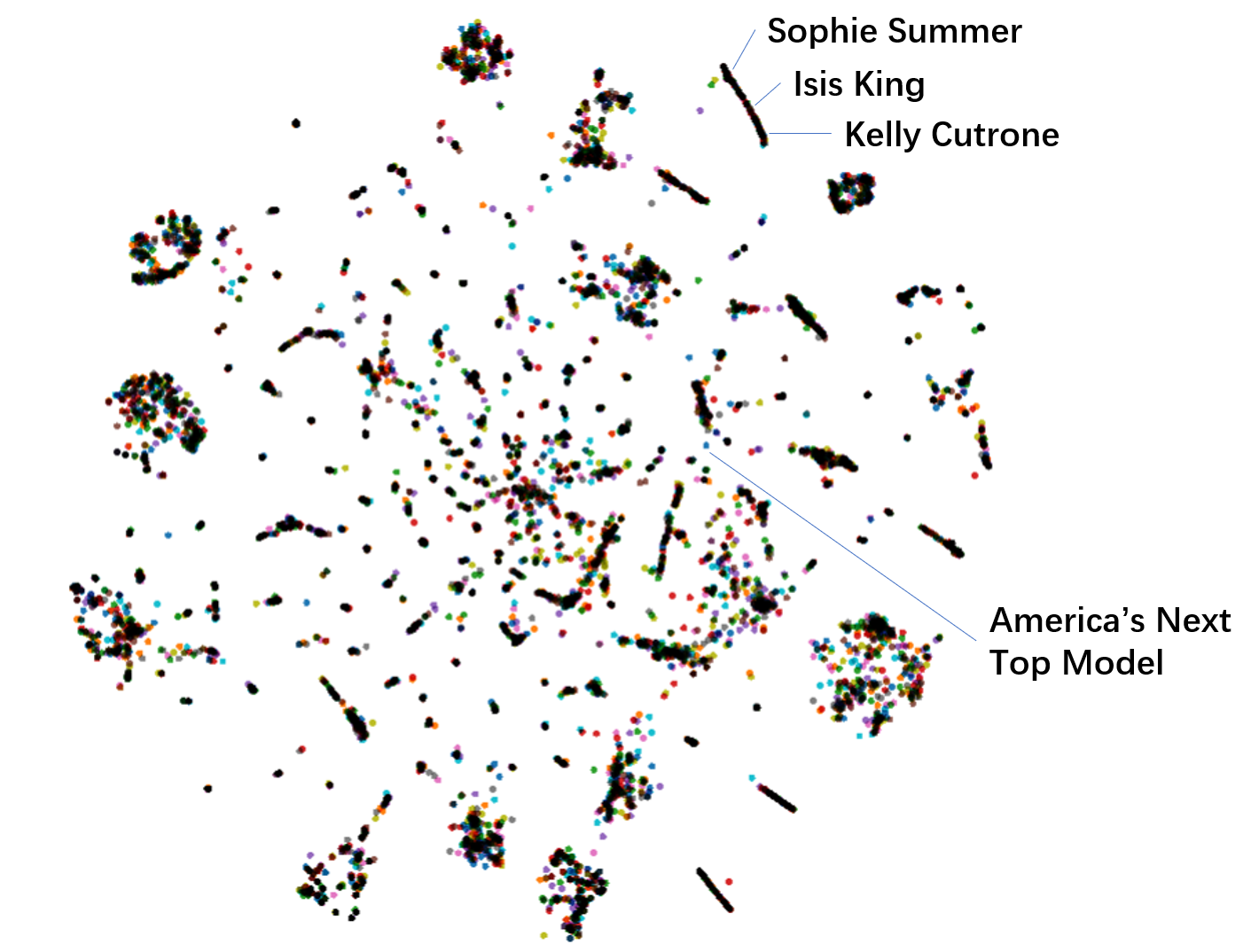}
		\label{fig_4_a}}
	\hfil
	\subfloat[Visualization of HINGE]{\includegraphics[width=3.3in]{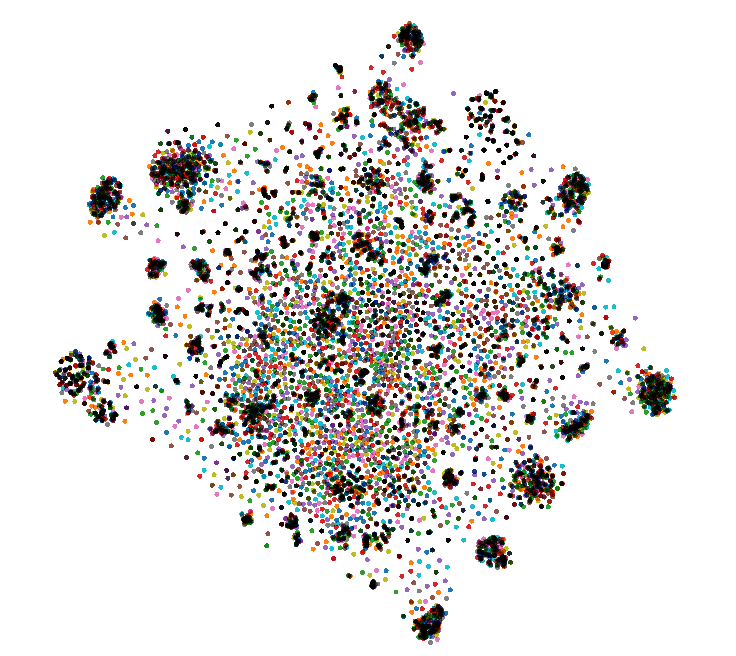}
		\label{fig_4_b}}
	\caption{In (a),  most entities are distributed near the border, few entities with high degrees lie more inward, which is in coherence with the nature of a Poincar\'e ball (see in Figure \ref{figure1}) where high-degree points are generally high in hierarchy, few in number and lie closer to the origin. In (b), large quantities of entities are distributed randomly. Intuitively, clusters in (a) are distributed more dispersedly and the area of a single cluster is smaller in (a), which illustrates that the embeddings learned by HYPER\(^2\) has larger inter-cluster divergence and smaller intra-cluster divergence. i.e., HYPER\(^2\) is superior to HINGE.}
	\label{figure4}
\end{figure*}

\subsection{Case Study}
In this section, we visualize the 28645 entitis in JF17K in accordance with their respective distance to the origin. Figure \ref{figure4}(a) is the embedding learned by HYPER\(^2\), Figure \ref{figure4}(b) is the embedding learned by HINGE. As shown in Figure \ref{figure4}(a), take America's Next Top Model, which is an annual TV show that did not close dwon until 2018, as an instance. With lots of models involved in the show every year, the entity \emph{America's Next Top Model} is high in degree, thus lying closer to the origin. \emph{Sophie Summer, Isis King and Kelly Cutrone} are three models who ever took part in the show. The three entities has lower degrees, therefore lying closer to the border. It is understandable that those entities high in degree are prone to lie inward and those entities low in degree tend to lie more outward, which is in coherence with the nature of a Poincar\'e ball (see in Figure \ref{figure1}) where high-degree points are generally high in hierarchy, few in number and lie closer to the origin. As we can see in Figure \ref{figure4}(a), most entities are distributed near the border. while in Figure \ref{figure4}(b), large quantities of entities are centered around the origin and distribute more randomly. 

Moreover, clusters distribute more dispersedly in Figure \ref{figure4}(a) and the area of a single cluster is smaller in Figure \ref{figure4}(a), which demonstates that the embeddings learned by HYPER\(^2\) are in possession of larger inter-cluster divergence and smaller intra-cluster divergence. Besides, in Figure \ref{figure4}(a), it's seen that the three individuals lie close to each other in the same cluster, which could be explained by the fact that they share similarity in their career and they all ever took part in the same show.
From all the experimental results and the analysis above, it's safe to draw the conclusion that HYPER\(^2\) demonstrates superiority over the ever best-performing baseline HINGE.


\subsection{Complexity Study}
In this part, we will theoretically and experimentally compare the parameter size and time complexity of our proposed HYPER\(^2\) against a collection of previous best-performing baselines. Since NaLP-fix differs from NaLP in negative sampling strategy but shares the same structure, it is no longer taken into our comparison in Table 6.
 We abbreviate affiliated entity to A-Entity in the reported results. 

\subsubsection{Parameter Size Study}
Regarding to the models to be compared, apart from the parameters introduced by entities and relations to be embedded, HINGE contains 2 extra convolutional layers and 1 fully connected layer, NaLP and NaLP-fix both include 1 extra convolutional layer and 2 fully connected layers. While HYPER\(^2\) involves extra biases for head and tail entities and a relation-specific diagonal matrix. Table \ref{table6} lists the parameter size and space complexity of NaLP, HINGE and HYPER\(^2\). Given along with the table are the notations. 

\begin{table}[htbp]
	\setlength{\abovecaptionskip}{0.cm}
	\caption{Paremater size and Space complexity analysis. }
	\label{table6}
	\resizebox{8.85cm}{!}{
	\begin{tabular}{llllr}

		\multicolumn{5}{l}{\textbf{\bm{$n_e$} and \bm{$n_r$} denote the number of entities and relations,  \bm{$d_e$} and \bm{$d_r$} are  } } \\
		\multicolumn{5}{l}{\textbf{the dimensions of entity and relation, \bm{$n_f$} and \bm{$k$} stands for the number }}\\
		\multicolumn{5}{l}{\textbf{of filters and kernel size of a cnn layer, while \bm{$w_{fcn}$} denotes parameter }}\\
		\multicolumn{5}{l}{ \textbf{brought by a FCN layer, \bm{$b_h$} and \bm{$b_t$} in HYPER\(^2\) are bias for head  and  }}\\
		\multicolumn{5}{l}{ \textbf{tail entity, \bm{$R_h$} is a diagonal matrix.}}\\
		\midrule[1.4pt]
		model & & & Parameter size                                                                                                               & \multicolumn{1}{r}{Space complexity  }           \\
		\cmidrule{1-5}
		NaLP  & & & ${n_e}{d_e}$+${n_r}{d_r}$+${n_f}k$+${w_{fc{n_1}}}$+${w_{fc{n_2}}}$ & $O({n_e}{d_e}$+${n_r}{d_r})$ \\
		HINGE & & & ${n_e}{d_e}$+${n_r}{d_r}$+$n_{f_1}{k_1}$+$n_{f_2}{k_2}$+${w_{fcn}} $  & $O({n_e}{d_e}$+${n_r}{d_r})$ \\
		HYPER\(^2\) & & & ${n_e}{d_e}$+${n_r}{d_r}$+${b_h}$+${b_t}$+${R_h}$                                                                          & $O({n_e}{d_e}$+${n_r}{d_r})$ \\
		\bottomrule[1.5pt]
	\end{tabular}}
	
\end{table}
Since when a hyper-relatonal graph is big enough, the parameter size is determined by ${n_e}{d_e}$+ ${n_r}{d_r}$, hence the space complexity of these models all can be represented as $O({n_e}{d_e}$+${n_r}{d_r})$. The model requiring fewer dimensions for entity and relation can be more suitable for large-scale application.

\begin{table}[htbp]
	\centering
	\caption{MRR of HYPER\(^2\) with Different Dimensions}
	\label{table7}
	\begin{tabular}{lccc}
	\toprule[1.5pt]
		model            & Head/Tail & A-Entity & Relation \\
		\cmidrule{1-4}
	NaLP             & 0.313     & 0.510             & 0.825    \\
	NaLP-fix         & 0.346     & 0.532             & 0.853    \\
	HINGE            & 0.490     & 0.575             & 0.898    \\
		\hline
		\cmidrule{1-4}
		HYPER\(^2\) dim = 10  & 0.520     & 0.556             & 0.926    \\
		HYPER\(^2\) dim = 20  & 0.546     & 0.590             & 0.944    \\
		HYPER\(^2\) dim = 50  & \textbf{0.583}     & 0.623             & 0.949    \\
		HYPER\(^2\) dim = 100 & 0.578     & 0.629             & 0.954   \\
		HYPER\(^2\) dim = 200 & 0.579     & \textbf{0.635 }            & \textbf{0.958}   \\
	\bottomrule[1.5pt]
	\end{tabular}
\end{table}
Table \ref{table7} are results of HYPER\(^2\) with different dimensions on JF17K, the dimensions for NaLP, NaLP-fix and HINGE are set to 100. It is notable that HYPER\(^2\) with 10 dimensions can outperform all of its counterparts with regard to head/tail entity prediction and relation predition. While for affiliated entity prediction, HYPER\(^2\) can also exceed HINGE and other baselines with only 20 dimensions. The results illustrate the superiority of HYPER\(^2\) in terms of parameter size as well as the potential of HYPER\(^2\) for large-scale application. We also note that more dimensions don't necessarily bring constant perfermance gain, fluctuation can be observed in head/tail prediction when we set dimension of HYPER\(^2\) from 50 to 100 or to 200.

\subsubsection{Time Complextity Study}
Theoretically speaking, the time complexity for a convolution layer can be denoted as:
\[O({C_{in}} \times {C_{out}} \times L \times W \times {L_f} \times {W_f})\]
where $C_{in}$ is the input channel size, $C_{out}$ is the output channel size, i.e., the number of filters, $L$ and $W$ denotes the the length and width of input map respectively, $L_f $ and $W_f$ represents the length and width of the filter. While for a fully connected layer, the computational complexity is given by:
\[O(FC{N_{in}} \times FC{N_{out}})\]
where $FCN_{in}$ and $FCN_{out}$ stand for the input and output size of a fully connected layer. Compared to the complex convolution or the matrix multiplication in fully connected layer, the mainly used M\"obius addition in HYPER\(^2\) is much more time-friendly. Further looking into the the total time $T$ for evaluation, according to the evaluation protocols introduced in Section 5.1.2, we have:
\begin{equation}
	\label{equation17}
	T = t \times \sum\limits_{i = 1}^{\|{TEST} \|} {A(fact(i)) \times {n_e}} 
\end{equation}
where $t$ is the time to feed forward one fact, $\|TEST\|$ is the number of test facts, $ A(\bullet)$ calculates the arity of a fact and $n_e$ is the number of entities in the graph.

 For fairness of comparison, we evaluate HYPER\(^2\) and the baselines with the same hardware, i.e., a RTX3080 GPU and an Intel 10700K CPU. We set the dimension of HYPER\(^2\) to 100, in line with that of the baselines. We experiment on JF17K only.
\begin{table}[htbp]
	\centering
	
	\caption{Time for a Model to Evaluate JF17K}
	\label{table8}
	\begin{tabular}{@{}lclc@{}}
		\toprule[1.25pt]
		model & NaLP & HINGE & HYPER\(^2\) \\
		\cmidrule{1-4}
		time/(s)  & 29616    & 36862     & 601  \\
		\bottomrule[1.25pt] 
	\end{tabular}

\end{table}

The time expense of each model is shown in Table \ref{table8}. As observed, it takes only 601 seconds for HYPER\(^2\) to evaluate the 24568 test facts in JF17K, while the time is 29616 and 36862 seconds for NaLP and HINGE, respectively. Our proposed HYPER\(^2\) is virtually 49 times quicker than NaLP and 61 times quicker than HINGE to evaluate JF17K, which means it take less time for HYPER\(^2\) to evaluate a fact, i.e, HYPER\(^2\) has a smaller $t$ in \eqref{equation17}. In practice, a smaller t means a quicker response, which is particularly important for large-scale application.

\subsection{Ablation Study}

In this part, we study the influence of different distance scoring functions and compare different information aggregation strategies on the tangent space as well. Experiment is conducted on JF17K dataset, the MRR of head/tail and affiliated entity prediction will be reported. 

\subsubsection{Distance Scoring Function}
We employ a bilinear scoring function where the head entity is multiplied by a diagonal relation-specific matrix \bm{$R_h$}, while a translational offset vector \bm{$r$} is employed to transform the tail entity. We try to seek the influence of \bm{$R_h$} and \bm{$r$} by changing the form of the distance scoring function.
\begin{table}[htbp]
	\centering
	\caption{MRR of HYPER\(^2\) with Different Distance Scoring Function}
	\label{table9}
	\begin{tabular}{lcc}
		\toprule[1.25pt]
		Scoring Function            & Head/Tail & A-Entity  \\
		\cmidrule{1-3}
		${d_\mathbb{P}}{(\bm{R_h}{ \otimes _K}{\mathbf{e}}_h^{hyper},{\mathbf{e}}_t^{hyper}{ \oplus _K}{\mathbf{r}})^2}$             & \textbf{0.583}     & \textbf{0.623}                 \\
		\cmidrule{1-3}
		
		$ {d_\mathbb{P}}{({\mathbf{e}}_h^{hyper},{\mathbf{e}}_t^{hyper}{ \oplus _K}{\mathbf{r}})^2}$        & 0.532     & 0.597                \\
		${d_\mathbb{P}}{(\bm{R_h}{ \otimes _K}{\mathbf{e}}_h^{hyper},{\mathbf{e}}_t^{hyper})^2}$            & 0.542     & 0.608                \\
		${d_\mathbb{P}}{({R_h}{ \otimes _K}{\mathbf{e}}_h^{hyper},{R_t}{ \otimes _K}{\mathbf{e}}_t^{hyper}{ \oplus _K}{\mathbf{r}})^2}$            & 0.535     & 0.601                \\
		${d_\mathbb{P}}{({\mathbf{e}}_h^{hyper}{ \oplus _K}{\mathbf{r}},\bm{R_t}{ \otimes _K}{\mathbf{e}}_t^{hyper})^2}$           & \underline{0.556}     &\underline{ 0.613}                \\
		
		\bottomrule[1.25pt]
	\end{tabular}
\end{table}

Table \ref{table9} shows MRR with different scoring function. As observed, any change to the bilinear scoring function would result in performance depreciation in varying degrees. Removal of relationa-specific matrix \bm{$R_h$} brings obvious degradation. Cancellation of translational offset vector $\bm{r}$ also leads to apparent debasement. Multiplying the tail entity with the realtion-specific diagonal matrix \bm{$R_t$} and translating the head entity with the offset vector $\bm{r}$ brings slight degradation, too.

\subsubsection{Information Aggregation}
 In our model, we design an information aggregator on the tangent space, which employs an element-wise minimization and an addition operation. We will experiment under addition and concatenation setting. Take the affiliated information specific head entity representation $\mathbf{e}_h^i$ as an instance, under the addition setting, the embedding of head entity and affiliated entity are added directly as what we do in \eqref{equation6}: \[{\mathbf{e}}_h^i = log_0^K({{\mathbf{e}}_h}) + log_0^K({{\mathbf{a}}_i}).\] While under the concatenation setting,  the embedding of head entity and affiliated entity are concantenated, i.e., $\mathbf{e}_h^i$ could be given as: \[{\mathbf{e}}_h^i = cat(log_0^K({{\mathbf{e}}_h}),log_0^K({{\mathbf{a}}_i})).\] Under each setting, aside from element-wise minimization, we try 
 element-wise maximization and average pooing as well. 
\begin{table}[htbp]
	\centering
	
	\caption{MRR of HYPER\(^2\) with Different Information Aggregation Strategies}
	\label{table10}
	\begin{tabular}{llcc}
		\toprule[1.25pt]
		\multicolumn{1}{l}{Setting}&\multicolumn{1}{l}{Aggregating Strategy} & Head/Tail & A-Entity \\
		\cmidrule{1-4}
		
		\multirow{3}{*}{addition}      & minimization     & \underline{0.583}             & \textbf{0.623}    \\
		& maximization     & 0.572            & 0.612    \\
		& average pooling     & 0.575             & \underline{0.614}    \\
		\cmidrule{1-4}
		\multirow{3}{*}{concatenation} & minimization     & \textbf{0.586}            & 0.471    \\
		& maximization     & 0.574             & 0.476    \\
		& average pooling    & 0.576             & 0.481    \\
		\bottomrule[1.25pt]
	\end{tabular}
	
\end{table}

MRR with different information aggregation strategies on JF17K dataset are reported in Table \ref{table10}. As seen, concatenating entity representation brings significant performance drop compared to original direct addition on affiliated entity prediction task, concatenation doesn't capture the interaction between primary entities and affiliated entities. We also note that element-wise minimization is the best information aggregation strategy under both settings, any change to the aggregation strategy would bring performance decrease in varying degrees, which is in coherence with our assumption that the framework telling right from wrong with least affiliated information is a better one.  

\section{Conclusion}
Knowledge Graphs are typically stored in triplet structrue, therefore existing Knowledge Graph embedding frameworks generally learn entity and relation embedding via triplets. Existing models seek to avoid tearing an n-ary fact into triplets, but still introduce quintuple-wise decomposition. In this work, we totally shrink from decomposition of an n-ary fact, keeping the integrity of a fact and maintain the primary triple. 
 Works on hyper-relational KG embedding are few in number and existing ones either suffer from weak expressiveness or high complexity in modelling n-ary facts. Under this circumstance, we propose HYPER\(^2\), a both effective and efficient hyperbolic poincar\'e embedding method for hyper-relational link prediction, which first maps entities and relations on a hyperbolic Poincar\'e ball and then aggregates the information from primary  and affiliated entities on the tangent space. Extensive experiments on two benchmark datasets (JF17K and WikiPeople) demonstrate the superiority of HYPER\(^2\) to its analogues, substantially outperforming its translational and deep frameworks. Besides, we further theoretically and experimentally prove that HYPER\(^2\) has an edge in parameter size and computational complexity. Moreover, we find literals in Wikipeople could impede the learning of a good embedding. As for future work, we plan to do more in-depth study in hyper-relational KG embedding by making allowance for entity type or multimodal data.


%

%
%


\bibliography{add}
\bibliographystyle{IEEEtran}
%

%

\begin{IEEEbiography}[{\includegraphics[width=1in,height=1.25in,clip,keepaspectratio]{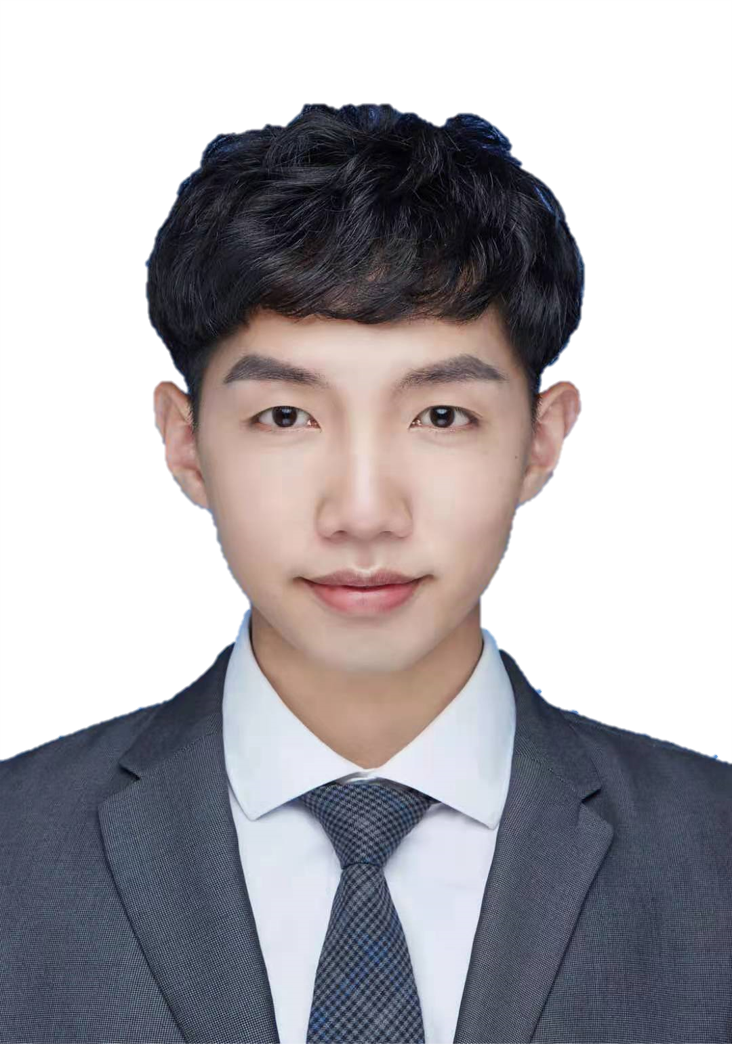}}]{Shiyao Yan} 
	
received the B.Sc. degree from Minzu University of China in 2019. He is working towards the PhD degree in Aerospace Information Research Institute, Chinese Academy of Sciences. His research interests include knowledge graph and data mining.
\end{IEEEbiography}

\begin{IEEEbiography}[{\includegraphics[width=1in,height=1.25in,clip,keepaspectratio]{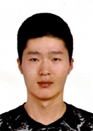}}]{Zequn Zhang}
received the B.Sc. degree from Peking University, Beijing, China, in 2012, and Ph.D. degree from Peking University in 2017. 
He is currently a Research Assistant at the Aerospace Information Innovation Institute, Chinese Academy of Science, Beijing, China. His research interests include information fusion, knowledge graph and natural language processing.
	
\end{IEEEbiography}

\begin{IEEEbiography}[{\includegraphics[width=1in,height=1.25in,clip,keepaspectratio]{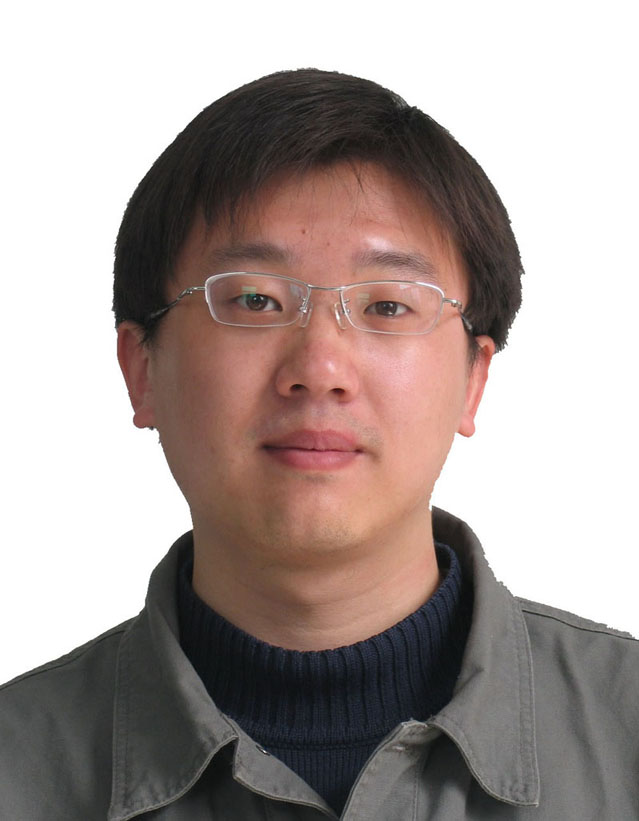}}]{Xian Sun}
received the B.Sc. degree from Beijing University of Aeronautics and Astronautics, Beijing, China, in 2004. He received the M.Sc. and Ph.D. degrees from the Institute of Electronics, Chinese Academy of Sciences, Beijing, China, in 2009. 
He is currently a Professor with the Aerospace Information Research Institute, Chinese Academy of Sciences, Beijing, China. His research interests include machine larning, computer vision, geospatial data mining.

\end{IEEEbiography}
\begin{IEEEbiography}[{\includegraphics[width=1in,height=1.25in,clip,keepaspectratio]{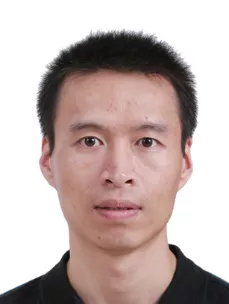}}]{Guangluan Xu}
received the B.Sc. degree from Beijing Information Science and Technology University, Beijing, China, in 2000, and the M.Sc. and Ph.D. degrees from the Institute of Electronics, Chinese Academy of Sciences, Beijing, China, in 2005. He is currently a Professor with the Aerospace Information Research Institute, Chinese Academy of Sciences, Beijing, China. His research interests include data minging and machine learning.	
\end{IEEEbiography}

\begin{IEEEbiography}[{\includegraphics[width=1in,height=1.25in,clip,keepaspectratio]{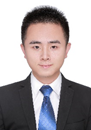}}]{Li Jin}
received the B.S. degree from Xidian University, Xi’an, China, in 2012, and the Ph.D. degree from the Department of Computer Science and Technology, Tsinghua University, Beijing, China, in 2017. 
He is currently a Research Assistant with the Aerospace Information Research Institute, Chinese Academy of Sciences. His research interests include machine learning, knowledge graph, and geographic information processing.
	
\end{IEEEbiography}
\begin{IEEEbiography}[{\includegraphics[width=1in,height=1.25in,clip,keepaspectratio]{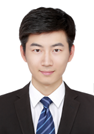}}]{Shuchao Li}
received the B.E. degree in software engineering from North China Electric Power University, Baoding, China, in 2012, and Master's degree in computer technology from North China Electric Power University, Beijing, China, in 2017.
Currently, he is a Research Assistent at the Aerospace Information Innovation Institute, Chinese Academy of Science, Beijing, China. His main research interests include knowledge graph, natural language processing and deep learning.
	
\end{IEEEbiography}




\copyright 2021 IEEE. Personal use of this material is permitted. Permission from IEEE must be obtained for all other uses, in any current or future media, including reprinting/ republishing this material for advertising or promotional purposes, creating new collective works, for resale or redistribution to servers, or reuse any copyrighted component of this work in other works.

\end{document}